*Article*

# Detecting Extratropical Cyclones of the Northern Hemisphere with Single Shot Detector

Minjing Shi [1] [2], Pengfei He [1]*  and Yuli Shi [1]*

1  Nanjing University of Information & Science Technology; miningshi@gmail.com
2  University of California San Diego; mishi@ucsd.edu
*  Correspondence: gopfhe@nuist.edu.cn; ylshi.nuist@gmail.com

**Abstract:** In this paper, we propose a deep learning-based model to detect extratropical cyclones (ETCs) of northern hemisphere, while developing a novel workflow of processing images and generating labels for ETCs. We first label the cyclone center by adapting an approach from Bonfanti et.al. [1] and set up criteria of labeling ETCs of three categories: developing, mature, and declining stages. We then propose a framework of labeling and preprocessing the images in our dataset. Once the images and labels are ready to serve as inputs, we create our object detection model named Single Shot Detector (SSD) to fit the format of our dataset. We train and evaluate our model with our labeled dataset on two settings (binary and multiclass classifications), while keeping a record of the results. Finally, we achieved relatively high performance with detecting ETCs of mature stage (mean Average Precision is 86.64%), and an acceptable result for detecting ETCs of all three categories (mean Average Precision 79.34%). We conclude that the single-shot detector model can succeed in detecting ETCs of different stages, and it has demonstrated great potential in the future applications of ETC detection in other relevant settings.

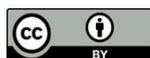

## 1. Introduction

*1.1 Background*

Extratropical cyclones (ETCs) are a common set large scale meteorological systems that are often formed in the middle to high latitudes for each hemisphere and feature intense variation in several horizontal and vertical physical quantities, such as temperature, air pressure, wind speed etc. [2] Because of the ETCs' huge influence on the ecological environment in terms of disaster weathers such as low temperature, storm in a wide span of regions, it is important for researchers and forecasters to automatically detect and identify them from satellite imagery to closely examine their characteristics since the images can bring instant and visual observations of the ETCs.

However, the automatic detection and identification of ETCs from images have been a difficult problem for a long time due to various unique features of the cyclones. First, any cyclones can bear varying degrees of mobility based on their size and geographic location. This makes it difficult to track down the cyclones by purely capturing the place at which the cyclones formed. Second, ETC systems bear large amounts of internal variations between individual samples. Some ETCs are small and regular-looking, while other ETCs can be gigantic in size and holding irregular shapes and features. Third, analyzing a sizable amount of ETC systems from a database generated by satellites requires a large amount of computing power, while also calling for a model complex enough to handle large variations between different samples. Finally, while the current several meteorological databases provide a plethora of ETC examples within the satellite images, they all suffer from a lack of clearly defined cyclone labels of ETC systems. These difficulties above have posed a significant challenge for mathematicians and meteorologists alike in the recent years.

Thankfully, the field of machine learning (ML), especially computer vision, have evolved a great deal over the past few years to be ready to handle complex and large-scale problems as such. There has been a surge of computer vision applications on sematic segmentation, object recognition and detection over the last few years due to numerous advancements in deep learning and artificial intelligence, as traditional computer vision methods such as scale-invariant feature transform (SIFT) descriptors [3] and Lucas-Kanade algorithms [4] are being replaced by more recent models that are directly adapted from deep convolutional neural networks such as AlexNet and ResNet [5, 6]. Such advancements are especially present in the field of object detection, where common models such as Region-based Convolutional Neural Network (R-CNN) and Object Detection called Single Shot Detector (SSD) are based on convolutional neural networks' various properties: they learn object representations at different scales of the images, and they are often robust to some level of noise in individual samples for datasets such as slight distortions, color shifting, and minor rotation[7, 8, 9, 10]. Because of this, object detection models are capable of handling many difficulties mentioned above, and they have great potential in tackling ETC detection and classification with image datasets.

*1.2 Recent works*

Currently, research that focuses on computer vision applications with respect to cyclone detection is somewhat lacking, especially in the category of ETCs, but there are several previous research that used either machine learning models or non-ML algorithms which lays the ground for the method proposed in this paper. One of the non-ML methods, proposed by Jaiswal, et.al., tackled the detection of centers of tropical cyclones through a model which utilizes gradient vectors of the brightness temperatures [11]. Another one written by Camargo, et.al. demonstrated a new algorithm that uses a grid to capture overlaps of several features of cyclones, including sea level pressure, relative vorticity, local temperature, etc., which facilitated tropical cyclone detection in atmospheric general circulation models [12]. While these non-ML methods are effective towards certain aspects of cyclone detection, for example, locating centers of tropical cyclones or finding a cursory location of the cyclones, they cannot pinpoint the exact area in which the cyclone systems span across, and often fails to achieve ideal results when the problem of cyclone detection is projected onto a global scale due to the lack of complexity and robustness to variations of samples for the algorithms.

On the other hand, cyclone detection research with machine learning algorithms works better when the amount of data is large regarding cyclones to process. Contrary to the smaller scale of non-ML algorithms employed to detect cyclones, they often cope with satellite imagery that captures the entire globe. Matsuoka, et.al. [13] adopted a deep learning approach to detect tropical cyclones and their precursors using two convolutional neural networks as classifiers and cloud images of long wave radiation generated from cloud atmospheric simulation as the inputs. At the same time, Kumler-Bonfanti et.al. [14] took the step even further by utilizing a semantic segmentation framework called U-Net [15] to segment the images collected from International Best Track Archive for Climate Stewardship (IBTrACS) [16, 17] and Geostationary Operational Environmental Satellite (GOES) [18], through the use of their custom labels which utilizes properties of cyclones such as vorticity, minimum sea level pressure, etc. Both approaches have been successful in locating and classifying tropical cyclone centers, with the latter research also drawing a boundary of the cyclone around the center with other clouds.

However, these works related to cyclone detection still have various downsides. The first problem is that they only attempt to detect the center part of cyclones, rather than the entire system, through the use of ML models. Besides that, few of them have established a general-purpose ML dataset that can be used interchangeably across all ML models, and their workflows regarding data processing and machine learning models are not transferable to other satellite image datasets. We would like to explore the possibilities of the detection of whole ETC systems through utilizing object detection models, rather than just detecting cyclone center parts. We would also like to address the problem of existing datasets by creating a novel procedure for preprocessing images to make a ETC dataset that is well-defined and versatile for object detection models. There are many other papers that frequently cites the term "cyclone detection", while the purpose of which are more geared towards forecasts of upcoming cyclones. For the sake of clarity, this paper would only be discussing cyclone detection with respect to computer vision techniques, using only imagery information.

In this paper, we attempt to address the problem of ETC detection of the northern hemisphere by applying object detection models in computer vision to a ETC dataset we created using images. Our paper has two specific aims: First, we would explore the ability of SSD model to detect whether there is a certain category of ETC systems present in the images, where datasets of ETCs with just one class is involved in training and evaluation. Second, we attempt to test the ability of the SSD model with multi-class detection and classification of ETC systems, where the model is given the dataset of ETCs of all three categories we labeled based on their development stages. The rest of our paper is organized as follows: the data section would cover the procedures for which we utilized to acquire our original images, label the ETC systems, and produce the ETC dataset; the method section first covers the model structure and rationales of SSD, then describes our framework to detect ETCs using the dataset we obtained; the experiment and result section records the experiment settings and the results obtained from ETC detection, each of which is divided into two parts based on our different goals of binary and multiclass detection; the discussion section first compares the test results of SSD in binary classification of ETCs for experiments on different categories, then it discusses the reasons behind results obtained from the SSD multiclass detection of ETCs and compares that with the experiments on binary classifications. Finally, we conclude that this research would serve as an early exploration for computer vision applications in machine learning, which demonstrates the immense power of neural networks models in capturing the complex characteristics of ETC systems.

Our contributions:

1. We proposed a novel procedure to create labeled ETC datasets that can be used by all object detection models.

2. We explored the performance of Single Shot Detector in ETC detection of the northern hemisphere by conducting experiments on binary ETC detection for dataset of single ETC categories and multiclass ETC detection for the entire dataset.

2. Data

*2.1 Data Acquisition*

For our problem setting of cyclone detection, we use ERA-Interim reanalysis data, top net thermal radiation (TTR), mean sea level pressure (MSLP), and vorticity, generated by the European Centre for Medium-Range Weather Forecasts (ECMWF), which attempts to emulate clouds, air pressure, vortex of the atmosphere [19]. To ensure the var-iability of the cyclones presented within the data, we selected TTR images across a time span of 2 years from 2017 to 2019, and each image are captured six hours apart. In this sense, we collected a total of 2920 images from this period, with each image's resolution set at 1440 by 721.

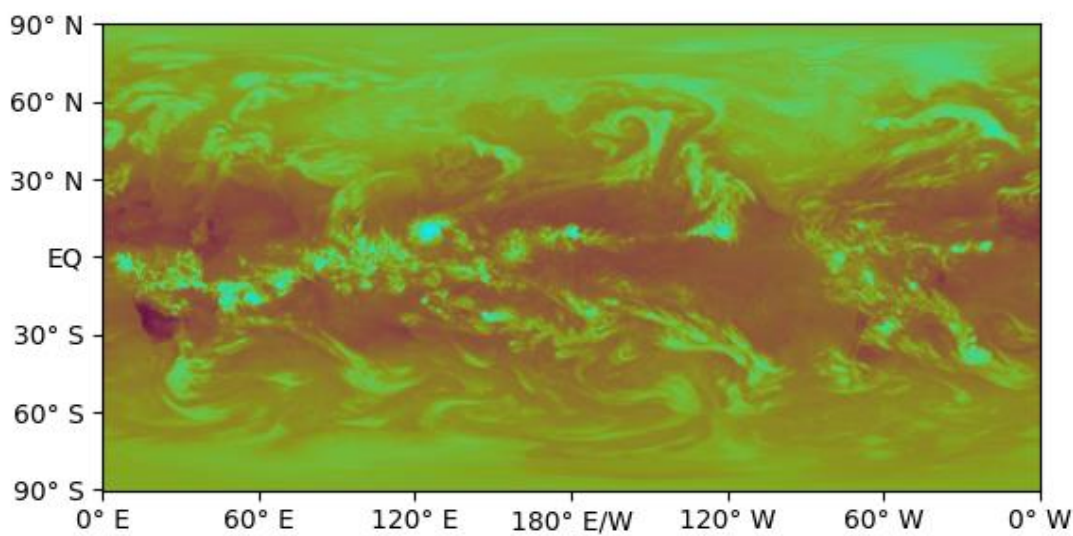

**Figure 1.** A typical image of top net thermal radiation (TTR) used in the processing of extratropical cyclone labeling and detection, which is from the ERA-Interim reanalysis dataset.

Figure 1 illustrates an example of the images in the data. As we can see, it captures a cloud simulation at a global level. There are several types of radiation data available within the dataset, including surface net thermal radiation, top net thermal radiation, and so on. We chose the TTR data because they simulate the satellite images in thermal band and cover the radiation both from clouds and surface objects, , thus making it possible to identify and classify cyclones. The other two types of data, MSLP and vorticity were also used to identify the center of ETCs.

*2.2 Defining Extratropical Cyclones*

Currently, there is few research that defines cyclones in terms of information obtained purely through images, with examples such as the cloud simulation input in Matsuoka.et, al.[13], and the dataset obtained through characteristics of cyclones for Bonfanti, et.al [1]. We define an ETC system through two steps: first, we locate the center of an ETC system and check the validity of this ETC system. Then, we divide the ETC into three categories according to characteristics of different development stages and make a bounding box around the center which is big enough to cover the cloud system this ETC belongs to.

To make sure our cyclones are well-defined bodies of meteorological phenomena, we adapted the approach taken by Bonfanti, et.al. [1], where they proposed a method for creating robust instances of cyclones through multiple frames of images captured by satellites. For our method of labeling unique ETCs, we take a slightly different approach that still centers around the idea of large distances between neighboring cyclones, and a relatively long time of existence. For the sake of this research, we only detect valid cyclones from the northern hemisphere. First, we pinpoint the locations of cyclone centers using an 8-point local minimum algorithm[1]. Then, we take 4 consecutive frames instead of 8 to represent a continued period of 24 hours, since our figures are obtained in 6-hour windows. The rest of the criteria for determining the displacements and neighbors for each cyclone are adapted from Bonfanti, et.al.[1] Namely, the cyclone should be dis-placed no more than 333.36 km per time step, and neighboring cyclones should bear a distance of more than 10 degrees in order not to be marked as the same cyclone. Hence, we have a brief summarization of the characteristics of cyclones defined in table 1. Based on these criteria, we apply a custom depth-first search function on the set of all the ETC centers found, so that we can find valid ETC centers that existed for over 4 consecutive frames and obtain an initial set of ETC systems that meets all the requirements.

Table 1. Summarization of the characteristics of cyclones center [1]

| TimeStep | Duration | Displacements between timestep | Distance from neighboring cyclones | Mean Sea level pressure |
|---|---|---|---|---|
| 6 hours | 24 hours, or 6 timesteps | <= 333.36km | >= 10 degrees | Local minimum (8 points) |

Moreover, we divided all valid cyclones into three categories, each representing a different stage of ETC: Developing stage (Figure 2a), Mature stage (Figure 2b), and Declining stage (Figure 2c). From the cyclone center, the experts label the associated cloud systems of ETC and mark its evolution stages. According to the conceptual models from ZAMG [20], a typical cold front cloud systems of an ETC often are found on the south side of the system during the development stage, and they appear as cyclonically curved large scale cloud bands and show white to grey colors in IR images. A large area of clear sky can be seen behind the cold front cloud system. A warm front shield or band lies in front of a cold front which is also at a synoptic scale in the image. A sharp gradient along the jet axis also cloud be found at the leading edge of a warm front cloud system. During the mature stage, the image shows a cyclonically curved cloud spiral at the synoptic scale. Meanwhile, the characteristics of the cold front and warm front cloud system are maintained or strengthened, and the occlusion front system reach to its maximum intensity and area. At the decay stage, the cloud spirals remarkably weaken, and the frontal clouds begin to break and dissolve. The experts removed tropical cyclones and only keep extratropical cyclone cases when labeling. They draw bounding boxes over TTR images, outlining the whole cloud systems of an ETC, include the cold front, warm front, and occlusion cloud system when they are present in an ETC system.

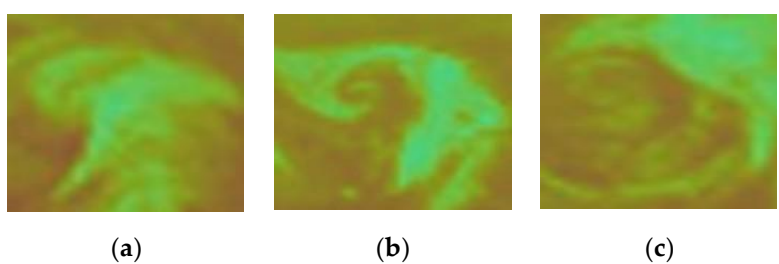

(a)  (b)  (c)

**Figure 2.** Typical images of three evolution stages of an ETC: (a) developing stage captured at 12:00 on March 4th, 2017; (b) mature stage captured at 20:00 on March 4th, 2017; and (c) decay stage captured at 4:00 on March 5th, 2017.

To ensure the quality of the ETC labels, we would have another two experts that check the labeling results. The first expert would review the labels and make suggestions based on the validity of ETCs and the label's boundaries, while the second expert ensures that these suggestions are backed by sufficient reasons. If there are any disagreements between the experts in the process of checking, both experts must come to a consensus through discussions over the specific labels. After extensive checking and discussion, we obtained labels of ETCs in the northern hemisphere for the year 2017 and 2018. As a result of extensive search and labeling, we collected a total of 1,507 valid ETCs of different stages. The shortest span of which come to 24 hours, while the longest of which come to roughly over 216 hours.

## 3. Method

*3.1 Model Selection*

To tackle the problem of ETC detection and classification, we turn to adapt a machine learning model that serves the purpose of detecting objects at multiple scales, while maintaining high accuracy with regard to multiclass classification. At the meantime, since the dataset we are dealing with is relatively large, with 2920 images in total, we also want the model to be easy to set up and takes less time to train and evaluate. In recent years, deep neural networks have gained widespread influence in the model design for object detection in computer vision. To achieve the goal of object detection, several models have been proposed in the last few years, with many of them based on two variants [7, 8, 21].

The first model variant, R-CNN [7], consists of a classifier network for regions in the image and another network that proposes bounding boxes in the image, with the two network first trained separately before being assembled. To detect objects in the image, the model utilizes both networks, with the region proposal network locating the objects, while the classifier network categorizing the objects in the proposed regions. Though R-CNN was able to achieve relatively higher performance compare with the other variant, this advantage comes at the cost of extremely high computing complexity and longer training time. R-CNN is also relatively difficult to set up due to the model being unable to obtain results directly from the input images, and that the model requires multiple steps in training and evaluation, occupying large amount of space in disks for the saved states. The second model variant, called single shot detector (SSD), attempts to both locate and classify objects in images with a single network. SSD holds several advantages when compared with traditional object detection models based on R-CNN. First, SSD is easier to train and evaluate because the model integrates both region proposal and classification, in which images are processed by the model in one pass instead of two. This greatly shortens the time required to train and evaluate the model. Second, SSD benefits from a relatively low model complexity due to reduced number of trainable parameters compared with R-CNN, making the model easier to operate on mobile devices such as smartphones, cameras, or drones.

As R-CNN's downsides cannot meet our criteria of model selection, we turn to use the SSD model [8]. Since the process of region proposal and classification is simultaneous, the model operates on a predefined set of templates, or ratioed rectangles, which represents the shape of the bounding boxes for potential objects in images. Images are first fed into the network and downscaled by a certain ratio after going through each hidden layer, and all these layers are evaluated separately using the arbitrary shapes of templates. Thus, each layer presents potential objects of interest at different scales, for which bounding box templates corresponding to that scale is proposed and drawn on the objects. To extract features and distinguish different classes of objects, SSD operates in a similar way to common convolutional neural networks such as VGG-16 (a backbone to SSD) [22], which attempts to record the patterns that represent objects of different classes and establish clear boundaries between classes through dimensional reduction, convolution, and backpropagation from loss functions. The loss for this model is evaluated on two aspects: confidence and localization. The formula is presented as follows [8]:

$$L = \frac{1}{n}(L_{conf} + \alpha L_{loc}) \quad (1)$$

where $L_{loc} = \sum_{pos}^{N} \sum_{bbox} x * smooth_{l1}(lb - gt)$,

and $L_{conf} = -\sum_{pos}^{N} x * \log(softmax(c^p)) - \sum_{neg} \log(softmax(c^n))$.

Confidence loss, or $L_{conf}$, indicates the level of truth the model computed for the class each detected object belonged to. This is computed by taking the negative SoftMax loss over multiple positive class confidences $c^p$ with input parameters $x$, and subtracted by the losses from the negative class confidences $c^n$. Localization loss, or $L_{loc}$, represents the accuracy of the location for drawn bounding boxes with respect to the ground truth bounding boxes. This is computed by taking a Smooth L1 loss [23] between predicted box $lb$ and ground truth box $gt$ with the input parameters. The localization loss can also be tuned by changing the weight $\alpha$. When considering the performance of SSD, the two aspects of the loss functions are combined and divided by the number of total boxes N where the ground truths match the predictions. The lower the total loss, the better the performance.

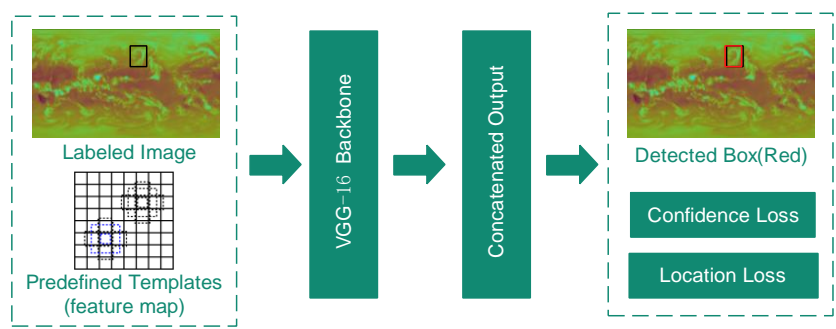

**Figure 3.** Structure of the Single Shot Detector Model.

*3.2 Overview of Workflow*

In this paper, we take the base SSD model proposed in the original paper regarding SSD, called SSD-300 [8], and utilize it for cyclone detection. we propose the following workflow for our experiment with SSD models. We first obtain the labels for all valid ETCs through the algorithms and labeling procedures described in section two, then we compile the labels of the dataset into a file and feed it along with the images into our object detection model for training. The images are preprocessed before being added into the model with several augmentations, including photometric distort, image translation to absolute coordinates, random sample crop, and random mirroring. The image is then resized to fit the model's requirements of images' input size, which is 300 by 300 pixels. As the image gets through each layer, its dimension shrinks, and templates that attempts to capture objects are applied accordingly in each layer to determine objects of various sizes. The final layer then makes a bounding box prediction and classification prediction for objects shown in the image. The combined loss for localization and confidence are then computed and propagated backwards to update the parameters in the model. When the loss for the model goes below certain threshold, we finish our steps for training and start evaluating the model.

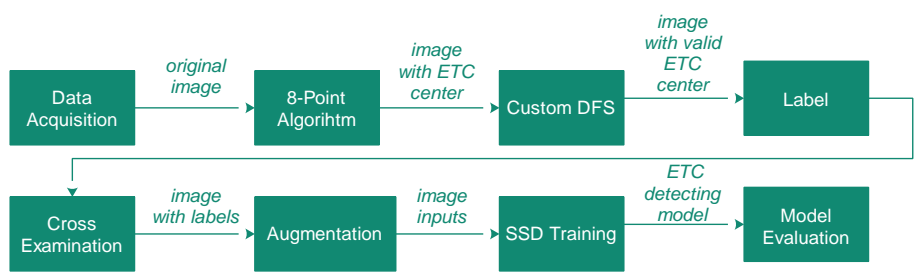

**Figure 4.** Workflow for our method.

To evaluate the performance of SSD, we use a metric called mean average precision (mAP), a metric commonly used in the object detection problem that measures the performance of relevant models. To compute this metric, we would first obtain precision and recall curves for the results of detecting each category in the dataset and compute the area under the precision-recall curve. This step gives us the average precision (AP) for the model in detecting this specific category. After we obtain the AP for all the categories in the dataset, we take the mean of these AP and compute the mean average precision, which reflects the model's performance in detection across all categories. For example, if we have three categories in our ETC dataset, we first compute the AP for each individual category (Developing, Mature, and Declining) in a multiclass detection evaluation, then we take the mean of these categories to compute the overall accuracy of the model.

## 4. Experiments and results

*4.1 Description of experiments*

In this paper, we are interested in exploring the performance of SSD models for ETC detection in various aspects: we want to evaluate the performance of SSD models on both binary (one vs none) and multiclass (one vs others) object detection problems. Hence, our experiments are divided into two parts. In the first part, we split our dataset into three groups: Developing, Mature, and Decline, according to the labels we defined in section 2. We would train an SSD on each of the individual dataset and check the performance of the model with the respective test set. In the second part, we consider the entire dataset for training and testing, hence the model is trained for multiclass detection of ETCs in all three stages. All results are evaluated with the respective test set and we use mean average precision (mAP) as the metric for model performance. We create specific test datasets corresponding to each sub-experiments, which are images selected at random making up 20% of the entire dataset. These experiments are run on a personal computer equipped with one CPU of Intel i9-10850k, 32gb of RAM, and one NVIDIA GeForce RTX 3080. For the SSD model we are using, we set the training iterations to 240,000, learning rate to $3 \times 10^{-4}$, batch size to 5, and the location loss factor alpha to 1. (Table 2 and Table 3)

*4.2 ETC Detection of Single Categories with SSD*

**Table 2.** Experiment for Binary ETC Detection for Dataset with Individual Categories.

| Category of ETC | Training Iterations | Training Time | Total Loss* | mAP | Training Samples | Testing Samples |
|---|---|---|---|---|---|---|
| Developing | 240,000 | 6 hours | 0.60 | 78.56% | 554 | 112 |
| Mature | 240,000 | 6 hours | 0.20 | 86.64% | 650 | 130 |
| Declining | 240,000 | 6 hours | 1.74 | 59.95% | 303 | 70 |

* Total loss indicates the loss computed in method section.

As we can see, SSD's accuracy in terms of detecting cyclones of a single class is the best for mature stage, topping at 86.64% in terms of mean average precision. The detection of Developing stage cyclones comes second, with a mean average precision of 78.56%. The lowest performance achieved for this experiment is detection on declining stage ETCs, only at 59.95%.

*4.3 ETC Detection of Multiple Categories with SSD*

**Table 3.** Experiment for Multiclass ETC Detection with the Entire Dataset.

| Training Iterations | Training Time | Total Loss* | mAP | Training Samples | Testing Samples |
|---|---|---|---|---|---|
| 120,000 | 4 hours | 1.39 | 69.65% | 1,507 | 300 |

| 240,000 | 6 hours | 0.59 | 79.34% | 1,507 | 300 |

* Total loss indicates the loss computed in method section.

Both training settings in this case achieved an accuracy of around 70 to 80 percent, with the model training at 240,000 epochs yielding a much higher accuracy. We tested different epochs of training time to see the effects on the accuracies with this variation. In this case, increasing the training time leads to a sizeable increase in accuracy.

## 5. Discussion

From the results in the experiment above, it is easy to notice that the SSD obtains the best result with detection (yes/no) of mature ETCs. This is because of all the valid ETC samples collected from our dataset, those that are labeled as mature stage ETC features the least variability between individual samples. All mature samples contain a visible spiral at the synoptic scale, as mentioned in the data section. They also feature a longer, more salient tail (of cold front), and resemble much like the number "9" when pieced together with the cloud spiral. Even though there are slight distortions for some samples of such shape, for example, tilts and squeezing of ETCs near the boundaries of the image for the northern hemisphere, the network can handle these variations between individual samples by learning the number "9" pattern (Figure 6). When the model is trained and tested with ETCs on the development stage, the accuracy dropped a bit from 86 percent to 78 percent, and loss grew from roughly 0.2 to 0.6. This can be explained with the following reason, that ETCs for development stage bears features that is both more variable and less pronounced than the mature ETCs. The ETCs of development stage feature slightly higher variations between individual samples compared with that from ETCs of mature stages due to the different factors in the formation of the system (Figure 5). The clouds would form different shapes at this early stage, with some samples caused by cold convey belt, while others affected by warm convey belt. Rather than featuring a visible spiral and a long tail, most of the system's clouds for developing ETCs are concentrated around the center with no obvious spirals, and the tail would often be too thin and short to be captured by the model. In a few instances, due to the model learning the less obvious features presented in development stage ETCs, it would mistake some larger tropical cyclones for ETCs of the development stage at test time, and it may ignore some smaller developing ETCs. In contrast to the higher performances of the model for the experiments with detecting Developing ETCs and mature ETCs, the model did poorly with the experiment on just the declining ETCs. This is mainly because during the declining stage, ETCs' structural integrity begin to break down. The strong spiral presented in mature stage starts to dissolve, and the clouds around the ETCs are also separating (Figure 7). At this stage, it is difficult for the model to recognize the entire structure of ETCs, since the strands of clouds which separated themselves from the ETC are often blended into the larger background with the cloud systems that are close to it. Therefore, the model often fails to recognize clear boundaries of cyclone systems with respect to the declining ETCs, and mistakes random cloud clusters for declining ETCs, resulting in the lower accuracies in the experiments compared with detecting developing and mature ETCs. Another problem may be that there are fewer samples collected for the declining ETCs compared with developing and mature ETCs, which may hurt the test accuracy due to the model's overfitting on samples from this category.

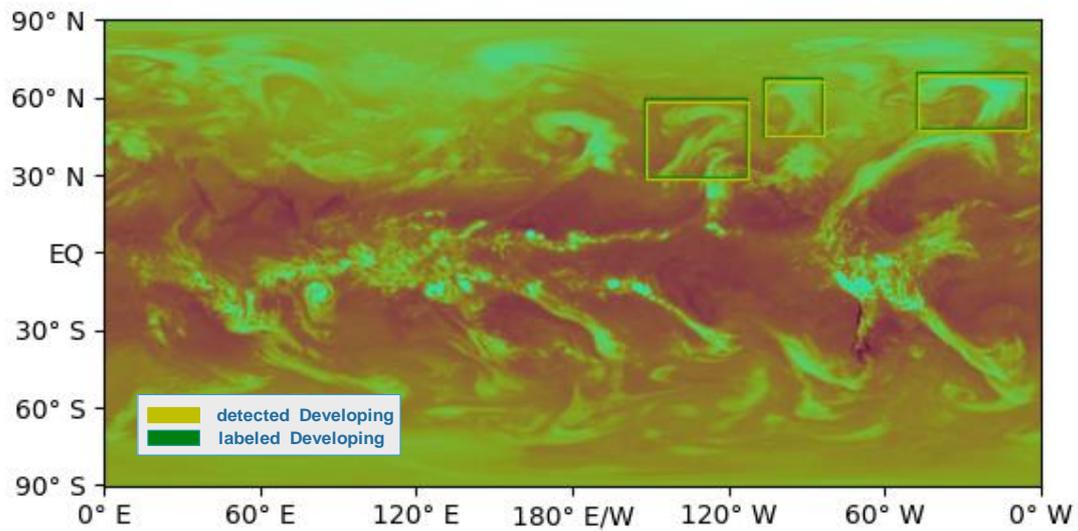

**Figure 5.** Correct testing examples of Developing stage ETCs. Some centers of ETCs are yet to be formed, and the ratio of the area of the center to the entire cyclones varies greatly. Still, they all exhibit relatively clear shapes and contours.

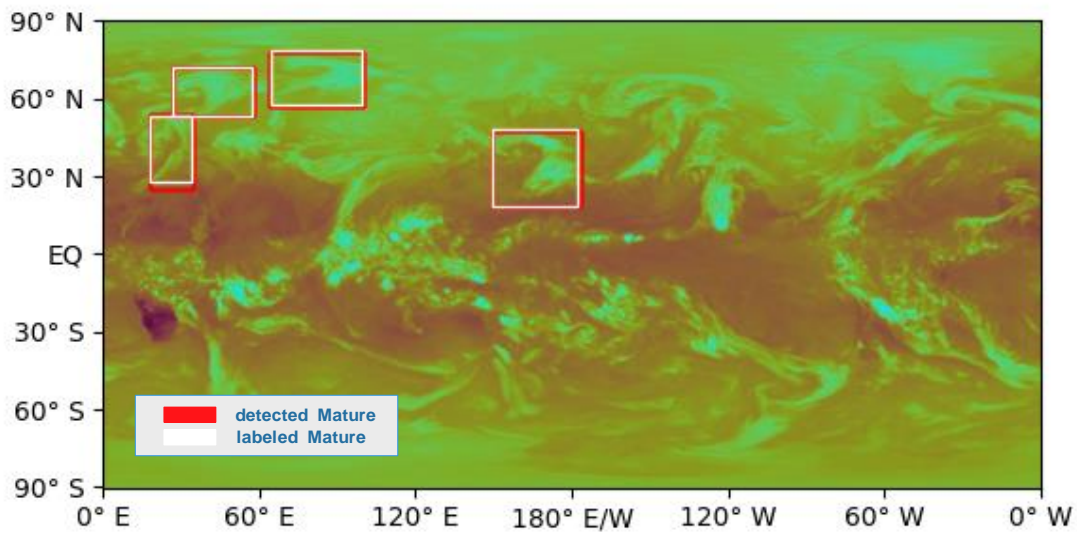

**Figure 6.** Correct testing examples of Mature stage ETCs. The samples all feature clearly visible spirals and tails, resembling the number "9".

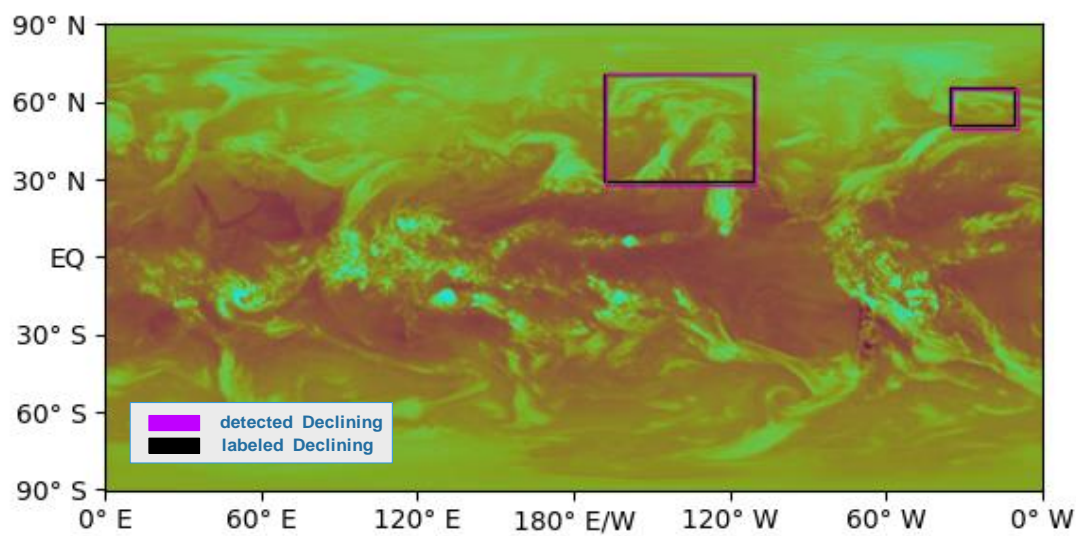

**Figure 7.** Correct testing examples for declining stage ETCs. There are few clear boundaries of cloud within the system that separates it from nearby clouds.

For the experiments with multiclass classification with all three ETC categories as an entire dataset, we saw a decreased performance compared with the experiments of datasets with individual categories of ETCs when trained with the same model and the same number of epochs. The results are most likely attributed to two reasons. One, there exists ambiguous labels within the dataset for which the cyclones are in transition stages; Two, for some ETC systems, it is hard to establish a clear boundary between them due to the embedding of one system into the other. Even though the samples for each category have been carefully selected, there still exists very few samples from different categories that resemble one another. Take the example of this picture (Number 2 in Figure 8). It is obvious that the picture contains ETCs of all three stages. While the model was right in locating and categorizing the ETCs of declining stage and mature stage, it somehow took the ETC of the developing stage as one that is in the mature stage. This is most likely because this particular sample of developing ETC was about to transition to a mature ETC, as the next picture shows (Figure 9), and this sample has the comparable size as that of a mature ETC. We can also notice that when the cyclone in (Number 2 Figure 9) was labeled as a mature ETC, there is a smaller ETC embedded into the center of this particular ETC (blue dotted rectangle in Figure 9). It is possible that the model mistook the first developing stage ETC because of the almost-developed center in (Figure 8), while in the next frame, this conception was enhanced due to the smaller ETC system resembling a strong cyclone center. This minor error creates some confusion between the model's learned classification boundary of ETCs of the different categories, resulting in mistakes for the classifier during testing and hence lowering the accuracy of the model for detecting ETCs. As we can see, the SSD model classifies objects in a similar way like common Convolutional neural networks through learning the different features existing between the categories of ETCs and tries to establish classification boundaries within these categories. When these boundaries of classification were distorted for the above reasons, the model would fail to correctly determine the type of cyclones existed in an image. For the tasks with identifying ETCs of individual categories, the model would suffer less since it is working on a rather smaller dataset with less variation between samples. This makes it easy for the model to learn the boundaries to classify objects, since there is only one type of template to establish for the SSD model and fewer outliers to avoid.

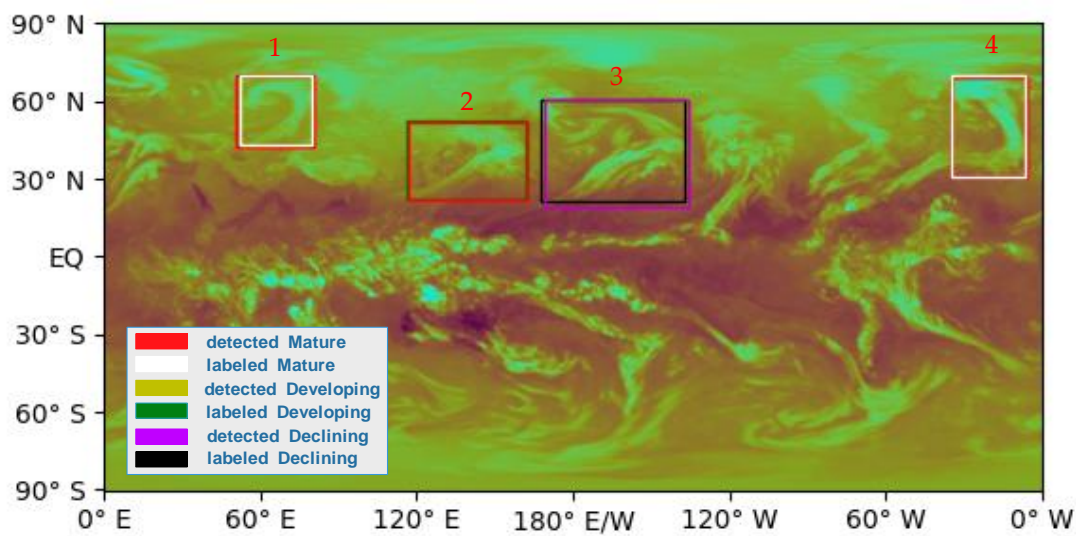

**Figure 8.** Example where a larger ETC at development stage is mistaken for mature stage. Numbers for bounding boxes are used to specify ETC examples in the discussion.

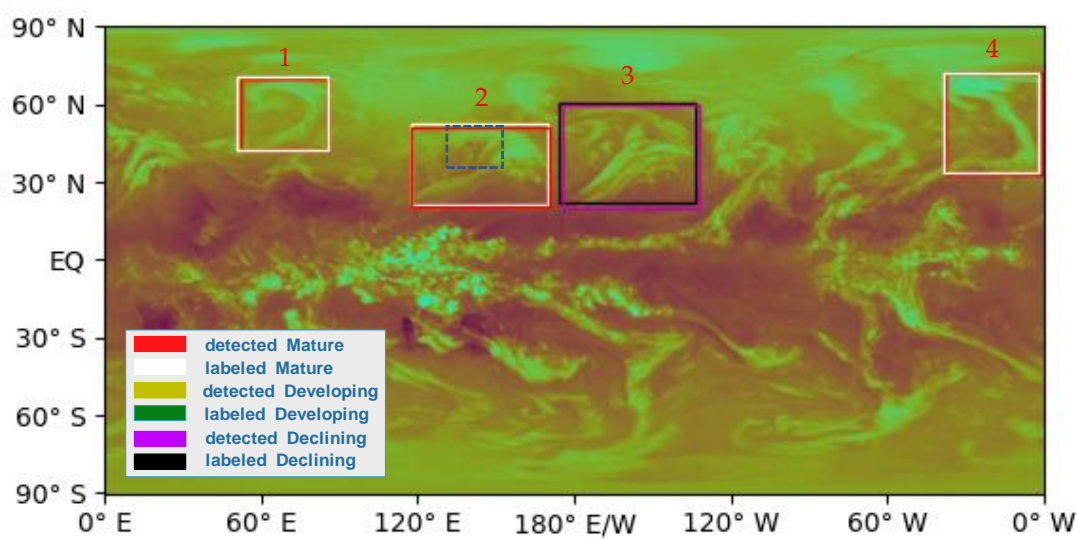

**Figure 9.** The next frame of figure 8. All samples are correctly predicted. Numbers for bounding boxes are used to specify ETC examples in the discussion. Blue dotted rectangle is a smaller ETC embedded into the center of a larger ETC.

## 6. Conclusion

In this paper, we proposed a novel framework to convert images of ETCs into labeled datasets, and we trained and tested the Single Shot detector for detecting ETCs of the northern hemisphere using our newly established dataset. We utilized algorithms such as DFS for data preprocessing through adopting terminologies proposed by Bonfanti, et.al. [1] and developing our own workflow of processing imaging data from the ECWMF. We categorized ETCs into developing, mature, and declining samples with careful hand labeling and cross examination between experts. We chose the Single Shot Detector as our model to train and evaluate, a model which learns features of different classes of ETCs and predict their locations on each image in one pass. We conducted experiments for the framework with datasets of individual categories and a dataset of all three categories combined, with the detection of mature ETCs yielding the best result among them all. In this way, we are successful with our aims since we have correctly set up a framework for training and evaluating SSD model on the ETC dataset, and we are able to achieve reasonable results for experiments on binary classification of ETCs and the multiclass experiment with ETC detection and classification. Going further from our current steps, the framework we implemented can be used directly in the study of satellite cloud images, and these results can be further optimized through fine tuning of SSD parameters.

As shown by our results, we believe that this research embodies some important explorations of applications of machine learning models in the field of remote sensing satellite imagery through attempts to establish a well-defined image dataset for cyclones, and a novel framework for processing such data and training such model. Eventually, we may develop a fully functional dataset used for weather object detection in remote sensing images just like MS-COCO [24] and VOC-2012 [25] for object detection in computer vision, and we will make modifications to the SSD and other object detection model so that they would be better suited to the object detection problems in special settings for remote sensing imagery. The field of machine learning, computer vision, and object detection have demonstrated their potential in the detection of everyday objects, and they may as well be powerful in settings relevant to remote sensing.

**Author Contributions:** Conceptualization, Yuli Shi; methodology, Minjing Shi; software, Minjing Shi; validation, Minjing Shi, Pengfei He and Yuli Shi; formal analysis, Minjing Shi; data curation, Minjing Shi; writing—original draft preparation, Minjing Shi; writing—review and editing, Pengfei He and Yuli Shi; visualization, Minjing Shi and Yuli Shi; supervision, Pengfei He and Yuli Shi; project administration, Pengfei He and Yuli Shi. All authors have read and agreed to the published version of the manuscript.

**Funding:** This research received no external funding.

**Conflicts of Interest:** The authors declare no conflict of interest.